\documentclass[letterpaper, 10 pt, conference]{ieeeconf}  

\IEEEoverridecommandlockouts                              

\usepackage{cite}
\usepackage{graphicx,subcaption}
\usepackage{float}
\usepackage{amsmath}
\usepackage{amssymb}
\usepackage{mathtools}
\usepackage{dsfont}
\let\labelindent\relax
\usepackage{enumitem}
\def\digits#1{%
  \number#1}
\usepackage{booktabs}
\usepackage{multirow}
\usepackage{pifont}%
\usepackage[dvipsnames]{xcolor}
\usepackage{pdfpages}
\usepackage{hyperref}

\newcommand{\tidyR}{\ensuremath{\hat{r}}}

\newcommand{\dataset}{\ensuremath{\mathcal{D}}}
\newcommand{\subfig}[1]{\textit{#1}}
\newcommand{\image}{\ensuremath{I}}

\newcommand{\object}{\ensuremath{O}}

\newcommand{\coordinate}[1]{\ensuremath{(x_{#1}, y_{#1}, z_{#1})}}
\newcommand{\optCoordinate}{\ensuremath{(x^*, y^*, z^*)}}

\newcommand{\hypo}[1]{\textit{\textbf{H#1}}}

\newcommand{\Semantic}{\emph{Semantic}}
\newcommand{\semantic}{\emph{semantic}}
\newcommand{\semantically}{\emph{semantically}}
\newcommand{\visual}{\emph{visual-spatial}}
\newcommand{\visually}{\emph{visual-spatially}}

\title{\LARGE \bf
``Tidy Up the Table'': Grounding Common-sense Objective for Tabletop Object Rearrangement
}

\author{\authorblockN{Yiqing Xu, David Hsu} \thanks{The authors are from National University of Singapore. Emails: \{xuyiqing,dyhsu\}@comp.nus.edu.sg. We thank Professor Leslie Pack Kaebling and Professor Tomas Loz\'{a}no-P\'{e}rez for their invaluable insights and discussions on the problem formulation and technical ideas.}
}

\begin{document}

\maketitle
\thispagestyle{empty}
\pagestyle{empty}
\begin{abstract} 
Tidying up a messy table may appear simple for humans, but articulating clear criteria for \textit{tidiness} is challenging due to the ambiguous nature of common sense reasoning. Large Language Models (LLMs) have proven capable of capturing common sense knowledge to reason over this vague concept of \textit{tidiness}. However, they alone may struggle with table tidying due to the limited grasp on the spatio-visual aspects of tidiness. In this work, we aim to ground the common-sense concept of \textit{tidiness} within the context of object arrangement. Our survey reveals that humans usually factorize \textit{tidiness} into \semantic{} and \visual{} tidiness; our grounding approach aligns with this decomposition. We connect a language-based policy generator with an image-based tidiness score function: the policy generator utilizes the LLM’s commonsense knowledge to cluster objects by their implicit types and functionalities for \semantic{} tidiness; meanwhile, the tidiness score function assesses the visual-spatial relations of the object to achieve \visual{} tidiness. Our tidiness score is trained using synthetic data generated cheaply from customized random walks, which inherently encode the order of tidiness, thereby bypassing the need for labor-intensive human demonstrations.  The simulated experiment shows that our approach successfully generates tidy arrangements, predominately in 2D, with potential for 3D stacking, for tables with various novel objects.
\end{abstract}

\section{Introduction}
Many common household tasks are governed by vaguely defined objectives rooted in human commonsense reasoning. Specifically, the task of tidying a messy table is guided by the somewhat vague concept of \textit{tidiness}: it's a concept readily understood by humans, but formulating a set of mathematically rigorous rules, reward functions, or constraints to quantify \textit{tidiness} can be challenging. Can we encode this vague concept of \textit{tidiness} into a robot, thereby enabling it to tidy up a messy table filled with various household objects? 

Grounding the common-sense concept of \textit{tidiness} on object arrangement presents significant challenges. The primary challenge stems from the need to consider a myriad of factors, such as the types, functions, visual attributes, and geometric properties of the objects. Grounding \textit{tidiness} necessitates reasoning over all these elements. Furthermore, these factors can sometimes introduce contradictions in the tidiness measure. For instance, grouping objects based on common usage might suggest placing a pencil near a notebook, while aligning objects for visual symmetry might advocate positioning a pencil next to a fork due to their similar appearance. As the variety and quantity of objects increase, the possible arrangements grow exponentially, adding to the complexity of the reasoning process.

To tackle this challenge, we conducted a survey \footnote{The complete survey and analysis can be found \href{https://drive.google.com/drive/folders/14io8VsNu6DuWx0HGyoWCHDvbxhXnLayN?usp=sharing}{here}.} to better understand how different factors influence the human perceived tidiness. The results indicated a hierarchy of importance among these factors, with over $90\%$ of participants prioritizing the grouping of objects of similar types and functions, followed by the visual neatness of the arrangement. Based on this insight, we decompose tidiness into two aspects: i) \semantic{}s tidiness, which pertains to the grouping and clustering of objects based on their semantic features such as types and functionalities, and ii) \visual{} tidiness, which assesses the visual appeal and geometric relations among objects. 

\begin{figure}[t]
\centering
\begin{subfigure}[t]{.32\linewidth}
    \centering\includegraphics[width=1\linewidth]{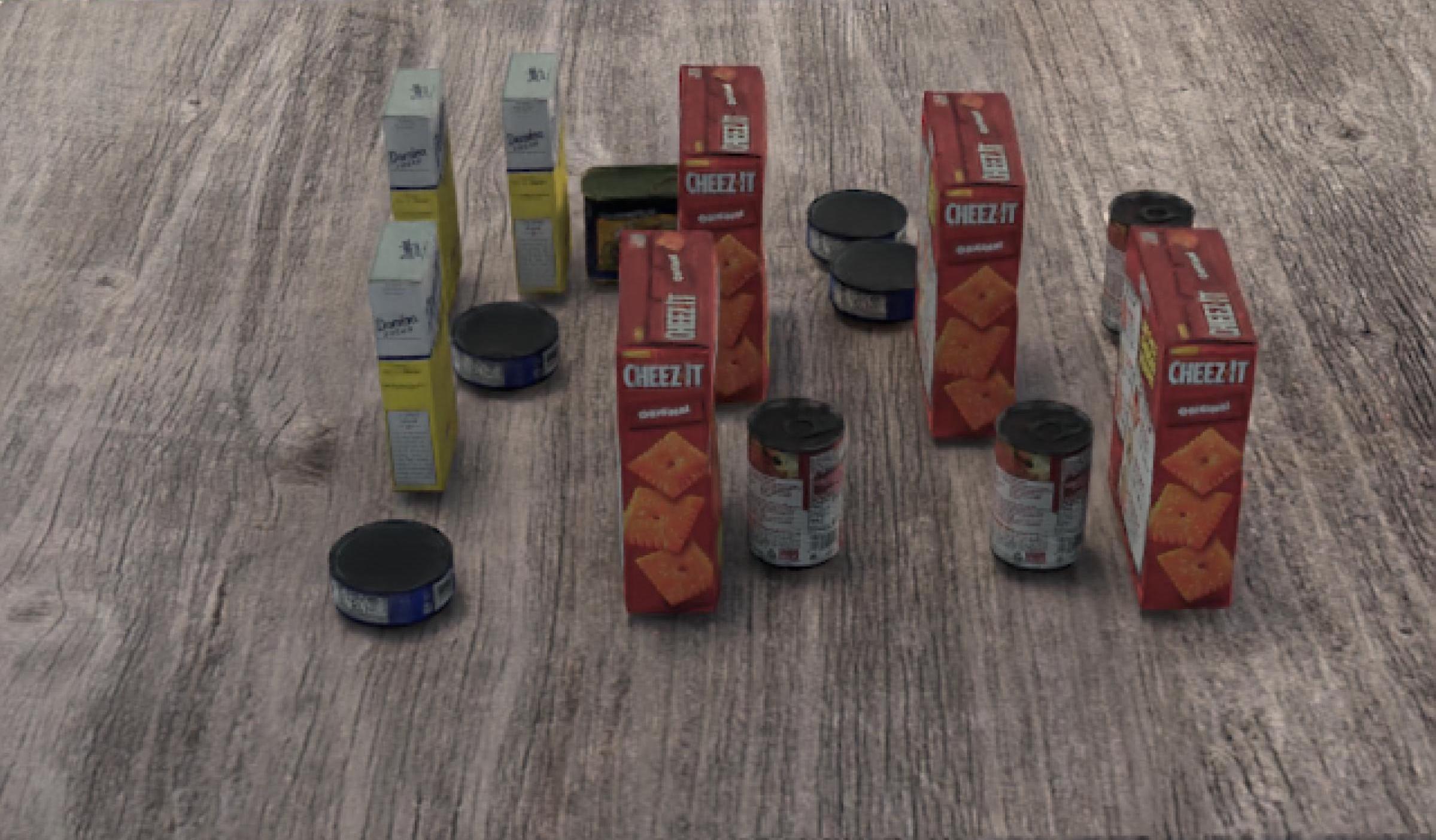}
    \caption{}
  \end{subfigure}
  \begin{subfigure}[t]{.32\linewidth}
    \centering\includegraphics[width=1\linewidth]{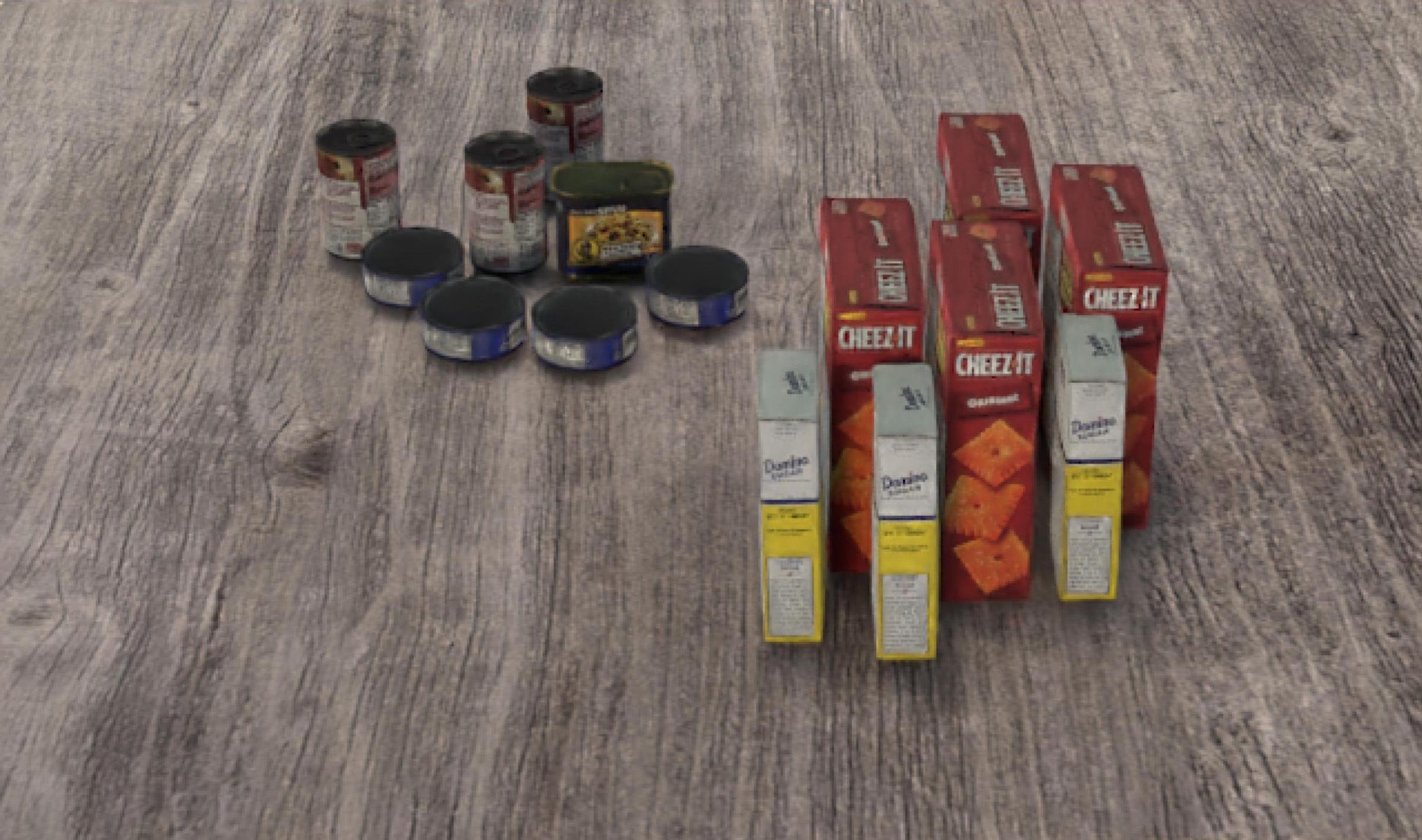}
    \caption{}
  \end{subfigure}
  \begin{subfigure}[t]{.32\linewidth}
    \centering\includegraphics[width=1\linewidth]{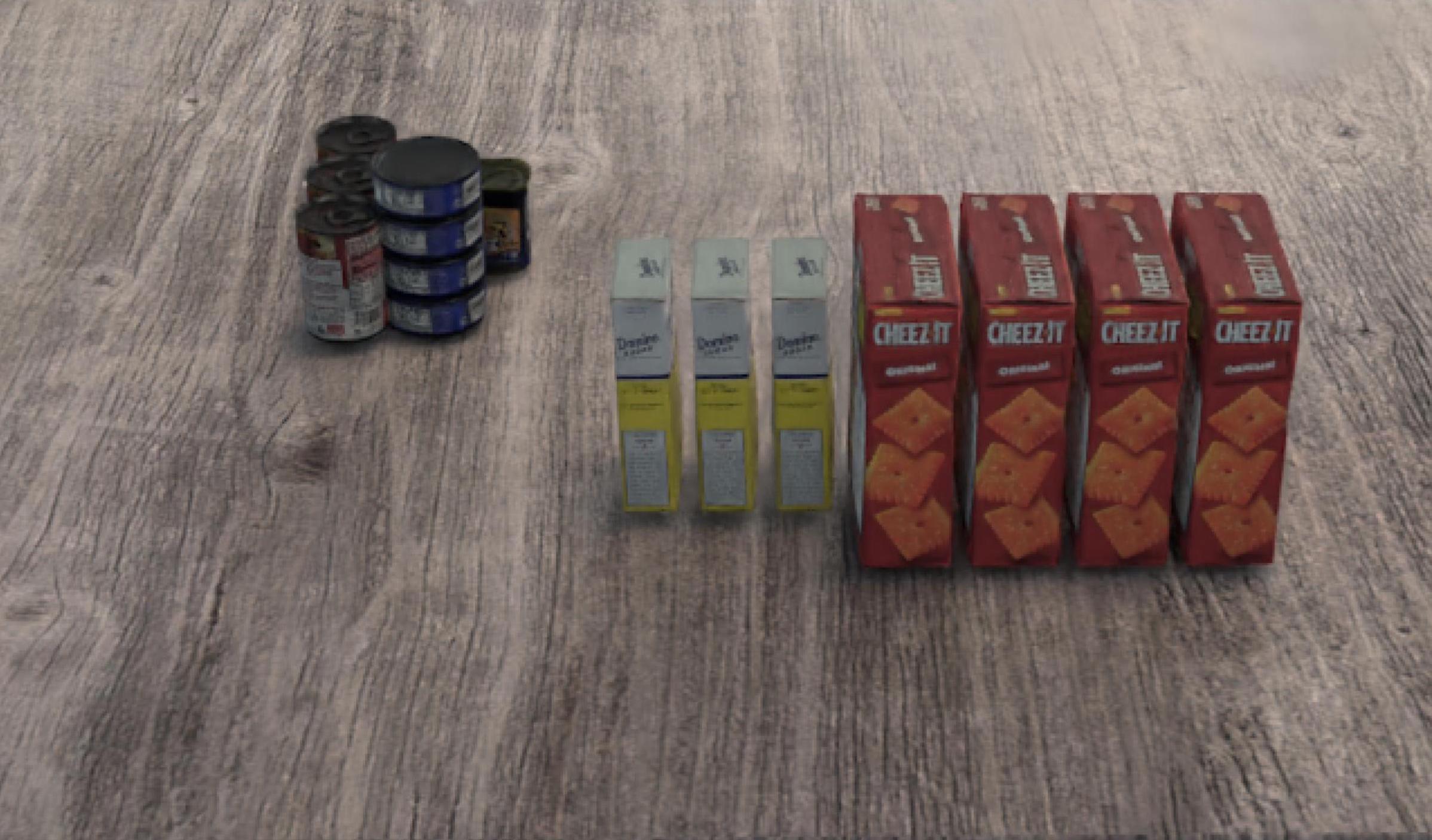}
    \caption{}
  \end{subfigure}
\caption{\small The criteria for a tidy table is vaguely encoded in common sense. 
To tidy up a messy table (\subfig a), we consider both grouping the objects based on their type/function to achieve \textit{\semantic{}} tidiness, as shown in (\subfig b), and arranging them to achieve visually pleasing arrangements (i.e. \textit{\visual{}} tidiness), as illustrated in (\subfig c).
}
\label{fig:tidiness_notion}
\vspace{-20pt}
\end{figure}

\begin{figure*}[ht]
\centering
    \includegraphics[width=1\linewidth]{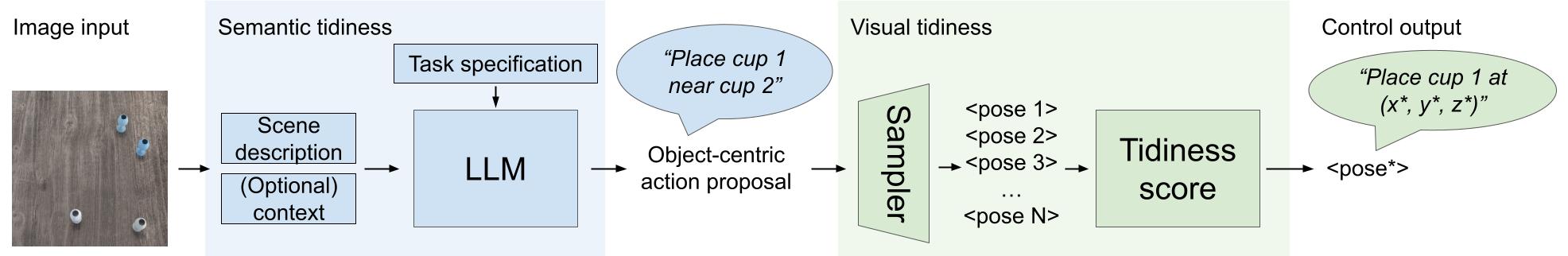} 
\caption{\small \textbf{The table tidying pipeline}, inspired by how humans factorize the concept of \emph{tidiness}, consists of two components: the LLM-based object-centric policy proposal, which groups objects based on \emph{semantic} features, and the image-based tidiness score, which grounds proposed actions to the \visually{} tidiest placements.}
\label{fig:pipeline}
\vspace{-20pt}
\end{figure*}

Our approach, illustrated in Fig. \ref{fig:pipeline}, mirrors this decomposition. It incorporates a pre-trained large language model (LLM) based policy generator that proposes object groupings based on \semantic{} features, followed by a trained image-based tidiness score that grounds the LLM-generated object placements into absolute coordinates for \visual{} tidiness. This pipeline aligns with the survey findings that \semantic{} tidiness precedes \visual{} tidiness. Through this decomposition, we factorize this complex problem into two simpler sub-problems that can be tackled independently.

Another challenge lies in creating a dataset that reflects the nuanced changes in tidiness across a broad spectrum of objects and arrangements to train the \visual{} tidiness score. Subtle visual variations in object arrangements can significantly change perceived tidiness. To effectively capture this, we need a training dataset that encapsulates gradually varying tidiness levels as the arrangements change. Binary datasets, which classify scenes as tidy or untidy, may not capture these gradual changes. While expert demonstrations can address this issue, the vast variety of object combinations and arrangements make achieving comprehensive data coverage impractical and costly.

\begin{figure}[t]
\centering
\vspace{7pt}
\begin{subfigure}[t]{.34\linewidth}
    \centering\includegraphics[width=1\linewidth]{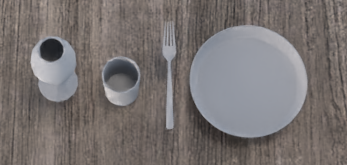}
    \caption{\small Known objects.}
  \end{subfigure}
  \begin{subfigure}[t]{.32\linewidth}
    \centering\includegraphics[width=1\linewidth]{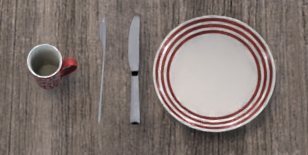}
    \caption{\small Similar objects. }
  \end{subfigure}
  \begin{subfigure}[t]{.295\linewidth}
    \centering\includegraphics[width=1\linewidth]{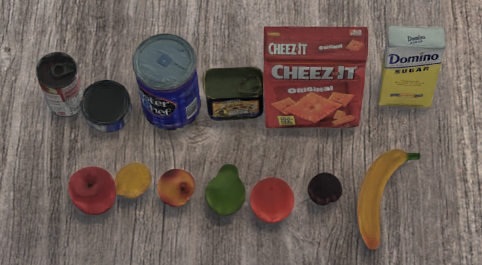}
    \caption{\small Novel objects.}
  \end{subfigure}
\caption{\small Objects used for training the tidiness score and evaluating generalizability are illustrated. Training data includes objects in (\subfig a), while evaluating generalizability includes some similar objects in (\subfig b) and the novel objects in (\subfig c). 
}
\label{fig:objects}
\vspace{-20pt}
\end{figure}

To address this, we exploit an intuitive ``entropy-based'' interpretation of tidiness to generate training data autonomously.  We assume that the tidiness represents a system's orderliness, and a random walk beginning from an orderly configuration incrementally introduces entropy and disorder. Drawing from this insight, we generate training data via random walks from various tidy configurations, randomly selecting and placing an object at each time step. The resulting sequences of object arrangements inherently encode a natural order of tidiness, which we subsequently utilize for preference learning. 

We conducted experiments on a wide variety of objects and layouts in simulation; the final tidiness performance was evaluated via a human study involving $30$ participants. The result shows that  our approach can effectively generate tidy arrangements, mainly in 2D with potential for 3D stacking, for tables filled with diverse unseen objects.

This study primarily focuses on grounding the common-sense concept  of \textit{tidiness}. We assume that objects are relatively sparse, hence forming object-centric manipulation policies based on object grouping is straightforward. However, largely cluttered settings pose considerable challenges due to partial visibility and complex task planning (e.g., placing one object near another requires decluttering to create more space first), which we plan to address in future work.

\section{Related work}
\subsection{Object Rearrangement}

Object rearrangement has received significant attention in recent studies \cite{simeonov2023se, zeng2021transporter, manuelli2019kpam, simeonov2021long, yuan2022sornet, qureshi2021nerp,driess2021learning, goodwin2022semantically, danielczuk2021object}. These studies primarily focus on deriving control policies for specified goal configurations, but they have not explored handling the vagueness in task objectives.  Some recent research has aimed to address ambiguities in language-based goal specifications \cite{zhao2023differentiable, liu2022structformer, liu2022structdiffusion, shridhar2022cliport, shridhar2023perceiver, lynch2023interactive, mees2022matters}. They specifically aim to ground natural-language instructions to precise object locations, which often involve spatial references (e.g., ``place the cup next to the plate'') or slightly more complex relational arrangements (e.g., ``rearrange the bowls in a line''). However, these approaches mainly deal with interpreting geometric and spatial interactions among objects and do not involve reasoning about the \emph{semantic} characteristics of the objects, unlike our approach.

Another line of works \cite{ batra2020rearrangement, szot2021habitat, weihs2021visual, puig2018virtualhome, li2021igibson,  gan2021threedworld} have focused on simulating household environments to explore tasks involving object rearrangement. These tasks require an agent to physically manipulate objects in order to achieve a specific pre-defined state. Unlike these tasks, the challenge of tidying up a table is unique due to the inherent ambiguity of the term \emph{tidiness}. This concept requires multifaceted reasoning that goes beyond spatial and geometric considerations.

\subsection{Large language models for robotics}
Large Language Models (LLMs), trained on extensive human data, have shown promise in generating task policies based on commonsense reasoning \cite{kojima2022large, nye2021show, rytting2021leveraging}. Several studies have utilized LLMs to enhance the autonomy of robotic systems
\cite{ahn2022can, brown2020language, huang2022language,  lin2023text2motion, wu2023tidybot}.  For example, in the context of table tidying, an LLM can propose logical object groupings based on human norms. However, directly implementing the LLM's suggested policy may not achieve the desired level of tidiness as they have limited ability to reason about visual and spatial details. While LLMs capture \semantic{} tidiness, \visual{} tidiness remains unaddressed.

Previous studies have explored different methods to ground the language-based policy from LLM to executable actions. Some approaches involve fine-tuning LLMs using human-labeled data to directly predict controls \cite{bakker2022fine, ouyang2022training, brohan2022rt, brohan2023rt}. Others define a discrete skill set for LLMs to plan over, training large visual models to ground each skill \cite{ahn2022can}, or assuming that skill execution aligns perfectly with task objectives \cite{huang2022language, wu2023tidybot}. However, obtaining human-labeled data is time-consuming and costly, and fine-tuning such large models can be prohibitively expensive. Furthermore, it might not be easy to design a set of skills capable of tidying up a table. Therefore, a cost-effective and flexible adaptor is needed to connect the \semantically{} tidy arrangements proposed by LLMs to \visually{} tidy placements.

\section{Proposed Approach}

\subsection{Problem Formulation}
Consider a table cluttered with various out-of-place household objects. Our objective is to re-arrange these objects into a layout that is both \semantically{} and \visually{} tidy. Specifically, 
we aim to produce a sequence of pick-and-place actions, denoted as  $\{\object_{i}, \coordinate{i}\}_{i=1}^T$, based on 2D images of the table layouts at each step, represented as $\{\image_{i}\}_{i=1}^{T}$. For each action  $\{\object, \coordinate{}\}$, $\object$ is the ID of the object being picked, and $\coordinate{}$ represents the placement coordinates. Executing these actions sequentially allows us not only to cluster objects according to their types, functions, and potentially some implicit semantic attributes for easy retrieval and usage, but also to place them in a visually pleasing arrangement.  We enlisted $30$ human participants to evaluate the final performance.

\subsection{Overview}
We decompose the common-sense concept of \emph{tidiness} into two aspects: \semantic{} and \visual{} tidiness, mirroring how humans typically perceive it. \Semantic{} tidiness assesses whether objects are reasonably grouped by their semantic features such as type and functionality, while \visual{} tidiness evaluates the orderly arrangement of objects through visual-spatial reasoning. Our approach, depicted in Fig. \ref{fig:pipeline}, merges a language-based policy generator with an image-based tidiness critic: the policy generator utilizes the LLM's commonsense knowledge to cluster objects by (implicit) type and functionality for \semantic{} tidiness; meanwhile, the tidiness critic promotes the \visual{} tidiness via visual-spatial reasoning over object arrangements.

\textbf{The language-based policy generator} takes the scene description and task specification and outputs a sequence of object-centric action proposals, such as ``\textit{Pick object\_1 and place it near object\_2}''. We also offer a few-shot variant that incorporates the user preferences by supplying $1$-$2$ sample solutions to the LLM. Crucially, the LLM-generated actions do not contain any absolute coordinates or poses, but rather focus on the desired clustering of the objects. 

\textbf{The image-based tidiness score function}, $\tidyR$, is used to ground the object-centric actions into exact coordinates that satisfy \visual{} tidiness. Specifically, this function compares the tidiness between two table configurations and outputs a score $\tidyR(\image_{a}, \image_{b}) \in [0, 1]$, representing the confidence that $\image_{a}$ is tidier than $\image_{b}$. To ground a language-based action, we first sample $N$ collision-free coordinates $\{\coordinate{0}, \ldots, \coordinate{N-1}\}$ that instantiate the object-centric action. Then, we form $N$ image pairs, each $(\image_{curr}, \image_{i})$ containing the current table image $\image_{curr}$ and the resulting image $\image_{i}$ after placing the object at the $i$-th sampled coordinate $\coordinate{i}$. The tidiness score function $\tidyR$ evaluates each of these pairs and the chosen coordinate maximizes the tidiness score:
\begin{equation}
    \optCoordinate \quad \text{s.t.} \quad \image^* = \underset{i\in [0, ..., N-1]}{\arg\min}\;\; \tidyR(\image_{curr}, \image_{i}) 
\end{equation}

\subsection{LLM-based policy generator}

\textbf{Scene Description} 
We use an off-the-shelf image segmentation model, Segment Anything \cite{kirillov2023segany}, to conceptualize the scene into a dictionary of objects and their semantic attributes. Trained on millions of images, it effectively segments and extracts semantic features for common household objects. We, then, convert the raw object dictionary to natural language description (via a template) to prompt the LLM for action proposals. As the LLM focuses on grouping objects based on their type and function, our template includes only minimal attributes relevant to \semantic{} tidiness.

\textbf{Prompting} 
Our LLM prompt has three components: scene description, task specification, and optional sample solutions. The scene description conceptualizes the table image, informing the LLM of all objects with attributes. The task specification outlines the goal (tidying up), table configuration, proposed action format, and the structure of the output format. Specifically, this fixed output format simplifies action proposal parsing and ensures commonsense consistency behind each action with a mandatory ``high-level organization rules'' following the ``instructions''. This becomes crucial when dealing with a large number of objects and longer task horizons. We also offer a few-shot variant to tailor action proposals to individual user preferences. By providing 1-2 sample solutions as demonstrations, the LLM, skilled at few-shot learning, adapts its proposals to align with demonstrated behaviors. \footnote{Access the complete set of prompts \href{https://drive.google.com/drive/folders/14io8VsNu6DuWx0HGyoWCHDvbxhXnLayN?usp=sharing}{here}.}

\subsection{Image-based critic: the tidiness score}

The image-based tidiness score $\tidyR$ captures the visual-spatial relations among objects to achieve \visual{} tidiness. 
Given pairwise data with labels on the relative \visual{} tidiness measure, we use preference learning \cite{christiano2017deep} to train the tidiness score on these synthetic data. To encode the visual-spatial object relations, we utilize a customized ResNet-18 to encode each image before inputting them into the final MLP to determine tidiness scores. 

\textbf{Preference learning}
We employ preference learning to train the tidiness function $\tidyR$, where a higher score indicates a tidier state. Intuitively, $\tidyR$ can be viewed as a predictor explaining implicit tidiness judgment, with the probability of a state being considered tidy exponentially depending on its tidiness score
\begin{equation}
    \hat{P}[\image_{t} \succ \image_{t'}] = \frac{\exp (\tidyR(\image_{t}))}{\exp (\tidyR(\image_{t})) + \exp (\tidyR(\image_{t'}))}.
\end{equation}
We minimize cross-entropy between these predictions and the labels of the training dataset, created using a customized random walk method detailed in section \ref{sec:random_walk}:
\begin{align}
     - \sum_{(\image_{t}, \image_{t'}) \in \dataset} & \mathds{1}_{ t < t'} \log  \hat{P}[\image_{t} \succ \image_{t'}] \nonumber \\ & + (1-\mathds{1}_{ t < t'})\log  \hat{P}[\image_{t'} \succ \image_{t}]
\end{align}

\textbf{Tidiness Network Architecture} 
The tidiness score network follows the Siamese network's structure \cite{chicco2021siamese} for pairwise data. Two images are first encoded using a customized ResNet-18 network \cite{he2016deep} with shared weights into two vectors. Our ResNet encoder has an additional channel that encodes the object segmentation maps of the scene. This is to inject the object-centric prior to train the tidiness score function. These vectors are then concatenated and passed to the final MLP layers to output the tidiness score. 

\subsection{Data collection}
\label{sec:random_walk}
The primary challenge of training the tidiness score lies in the data collection, as deducing the gradual change of tidiness between two consecutive images requires labels for all intermediate steps. To tackle this, we devise an efficient and cost-effective method to synthesize training data with natural labels using random walks. 

\textbf{Two-stage random walk}
Our tidiness score identifies the \visually{} tidiest coordinate among images with slight variations in a single object's placement. 
From our observations, it's clear that table tidiness decreases gradually if we initiate random actions from a few well-structured arrangements, akin to a random walk increasing system entropy. Accordingly, we constructed tidy configurations following the human preferences derived from the survey results. This enabled us to gather trajectories with natural order of tidiness without the need for human annotation.

We propose to generate the training data via a two-stage random walk. In the global stage, we start with tidy configurations and perform a random walk by randomly selecting and placing a single object at each step. In the local stage, we introduce multiple 1-step disturbances by slightly shifting the target object from each step of the global walk. This helps us collect more locally contrastive pairs. Images earlier in the trajectory have higher tidiness scores compared to later ones: for each image pair ($\image_{t}$, $\image_{t'}$) in a single random walk trajectory, $\image_{t} \succ \image_{t'}$ if $t < t'$. We can form pairwise preference orders using states in each trajectory and train the tidiness scores via preference learning. This approach offers several benefits. Firstly, it includes consecutive frames, allowing us to learn tidiness evolution over time. Moreover, the human effort is minimized: the random walk generates the tidiness ordering naturally, requiring human annotations only for the initial tidy configurations.

\textbf{Data selection}
We select preference image pairs from each trajectory to learn the pairwise tidiness score. We observe that a single random action can disrupt an initially tidy table; but as the table gets messier, the increase in disorder becomes noticeable only after multiple random actions. To address this, we generate more comparisons for early timesteps in the trajectories and gradually increase the intervals between comparisons as the timesteps progress, accounting for the nearly saturated disorder of the system.

\section{Experiment}
\label{section:experiment}
We conducted simulated experiments with randomly arranged messy tables, consisting of various types and numbers of objects arranged in different positions. We analyzed $100$ diverse table arrangements, wherein $70$ included similar objects present during the tidiness score training, with the remaining $30$ incorporating novel objects. 

Our experimental design intends to empirically assess three primary hypotheses:

\begin{enumerate}[label={\textit{\textbf{H\protect\digits{\theenumi}}.}}]
    \item An end-to-end policy proposal using LLM fails to ensure either \semantic{} or \visual{} tidiness.
    \item Decomposing the task into object-centric LLM policy proposal followed by generic action grounding (via collision checking) guarantees \semantic{} tidiness, but falls short on \visual{} tidiness.
    \item Training an image-based tidiness score can effectively ground the LLM's object-centric policy to achieve both \semantic{} and \visual{} tidiness. 
\end{enumerate}

We compare our approach against two baselines. The first employs an end-to-end LLM policy, prompting the LLM to generate action sequences with exact placements. We substantiate \hypo{1} by evaluating the final layouts of this baseline. The second baseline, similar in design to our approach, performs task decomposition: it prompts the LLM to exclusively propose the object-centric relative poses for placements, then grounds the proposed policy with collision checking. The final tidiness measures of the second baseline is used to validate \hypo{2}. Moreover, as the key distinction between our approach and the second baseline lies in action grounding, any performance discrepancy between the two serves as evidence to validate \hypo{3}. Lastly, we examine the generalizability of each method by tidying tables with larger and more diverse sets of unseen objects.

\begin{figure}[t]
\centering
\includegraphics[width=1\linewidth]{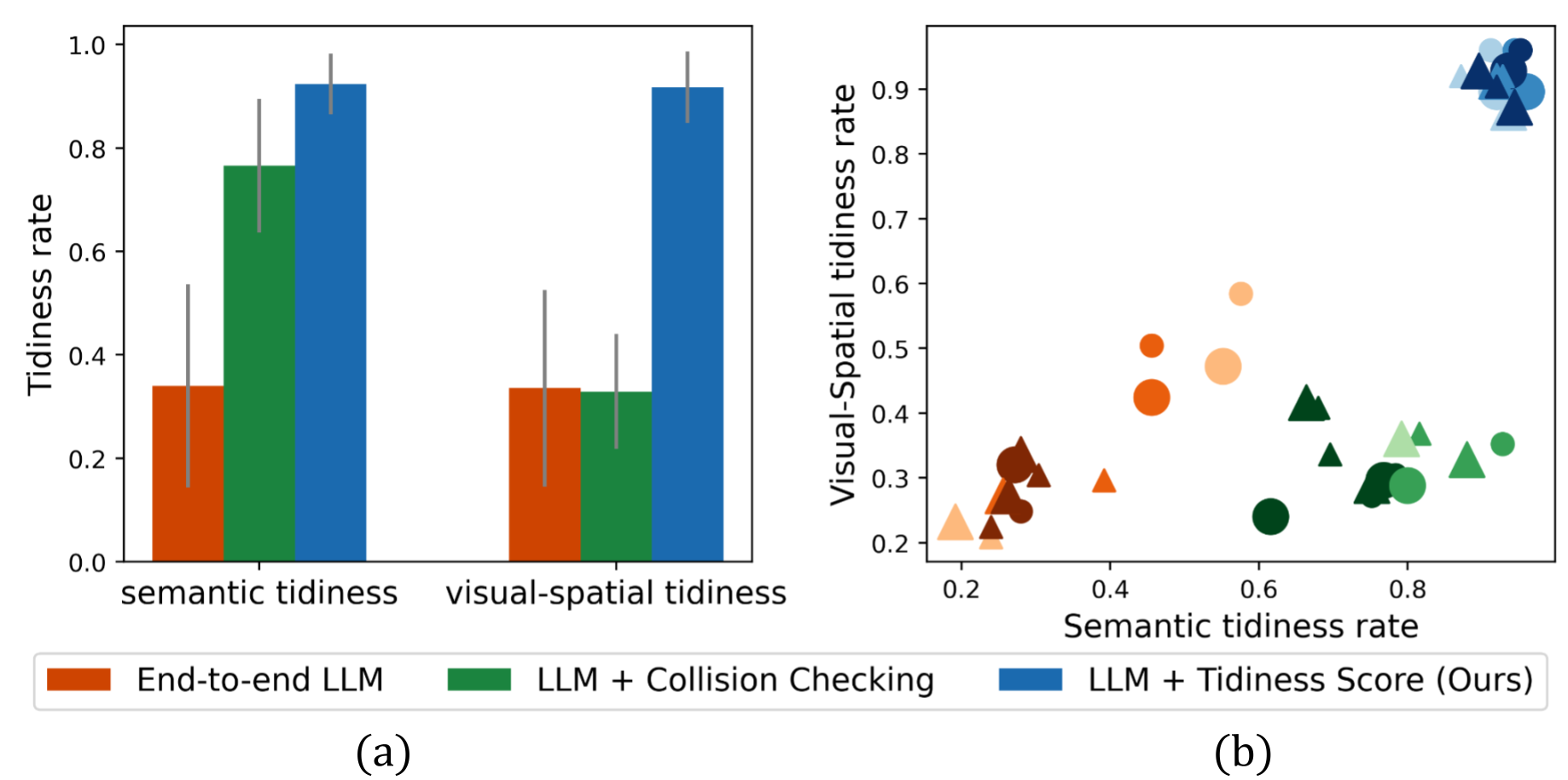}
\caption{\small \textbf{Comparison of Method Performance.} The bar chart in (a) displays average \semantic{} and \visual{} tidiness rates for each method across 70 test cases. Our method (blue bars) demonstrates high performance in both categories. The scatter plot in (b) showcases the correlation between tidiness rates and increasing table layout complexity for each method. Each marker represents a specific complexity level, defined by a combination of object types, quantities, and the requirement for 2D/3D spatial reasoning. Different color schemes indicate different methods, as illustrated in the legend. Each marker's color intensity and size are proportional to the object diversity and quantity, respectively. The $\triangle$ marker represents scenes requiring 3D stacking, while $\circ$ denotes 2D arrangements. The cluster of blue markers in the top right corner suggests our approach maintains \semantic{} and \visual{} tidiness despite layout complexity.
}
\label{fig:in_distribution_barplot}
\vspace{-20pt}
\end{figure}

\subsection{Experimental setup}

\begin{figure*}[ht]
\vspace{5pt}
\centering
    \includegraphics[width=0.85\linewidth]{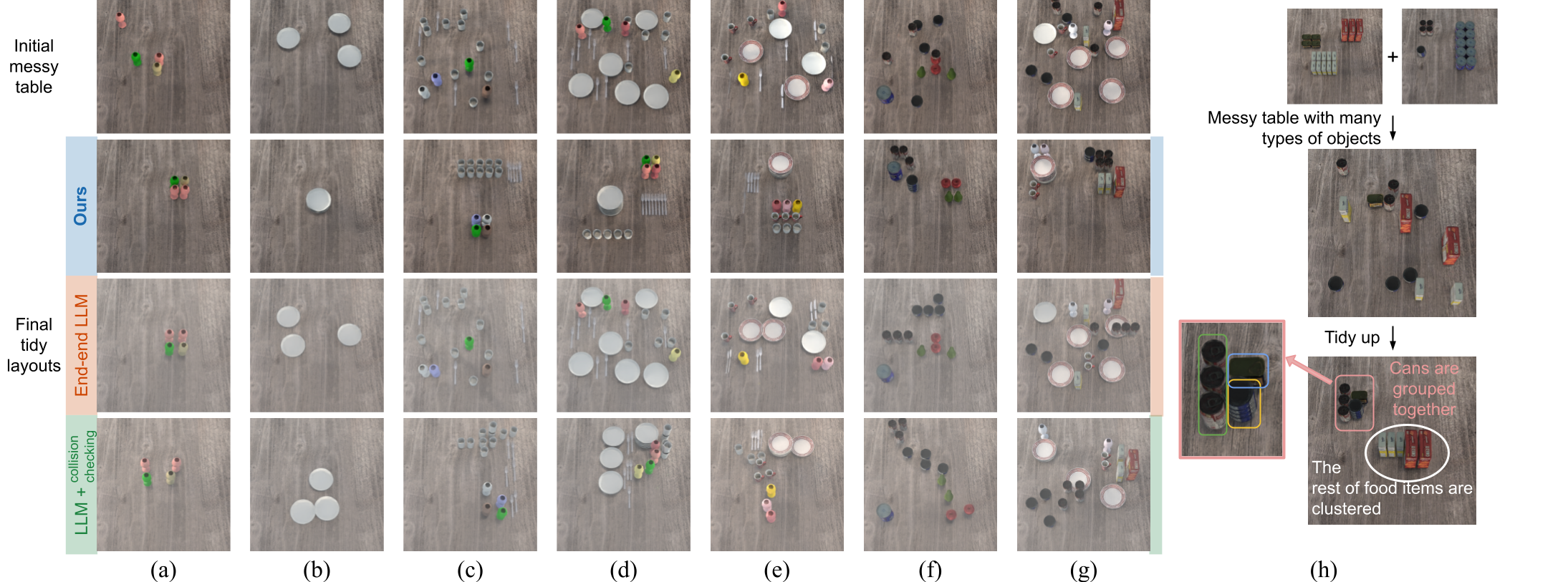} 
\caption{\small Highlighted instances from the experiment results, with each instance corresponding to a column. Col (a-d) depict instances with seen objects during the tidiness score training, while col (e-g) generalize to novel objects. The initial messy table arrangements are shown in the first row. We compare our method (4th row) with two baseline approaches (2nd and 3rd rows). (h) illustrates an challenging table setting featuring various object types, with few objects per type. Our pipeline divides objects into two groups and arranges them neatly. Notably, it organizes cans of different shapes, aligning tall ones (blue box), stacking thin ones (yellow box), and placing a single rectangular can in the remaining space (green box).}
\label{fig:tidy_instance}
\vspace{-20pt}
\end{figure*}

We use a high-fidelity tabletop simulator \cite{zhao2023differentiable} built on Pybullet \cite{coumanspybullet} as our simulated task environment. Each table setting involves objects shown in Fig. \ref{fig:objects}, with randomized combinations, quantities, and initial placements. Each pick-and-place action is defined by a tuple {$\object_{i}, \coordinate{i}$}, where $\object_{i}$ is the ID of the object to be picked and $\coordinate{i}$ signifies its precise placement coordinates. We used GPT4 \cite{openai2023gpt4} as the LLM for all experiments.

We evaluated the methods through a human study involving $N=30$ participants. Participants were presented with the initial messy table and the final layouts resulting from each method. They were asked to verify if each method achieved \semantic{} and \visual{} tidiness. A ``tidiness rate'' was calculated as the ratio of positive responses to total participants to each measure. Lastly, evaluators selected the tidiest arrangement from the three methods, with optional justification for their choice.

\subsection{Results}
We evaluated three methods across $70$ messy table layouts of varying complexity with in-distribution object types (those present during tidiness score training). The layouts varied from simple ones with a single object type, requiring only 2D rearrangement, to complex ones with multiple object types, many objects, and a need for 3D stacking. The evaluation result is summarized in Fig. \ref{fig:in_distribution_barplot}.

\textbf{Overall Performance} 
Our method consistently achieves over $90\%$ average rates in both \semantic{} and \visual{} tidiness. This indicates that our table tidying capability reaches human-acceptable level in nearly all scenarios, affirming \hypo{3}. Baseline 1 shows the least performance, not exceeding $32 \%$ for both tidiness rates, signifying the inability of LLM alone to tidy up the table, thus supporting \hypo{1}. Lastly, Baseline 2 presents a \semantic{} tidiness rate ($78.2\%$) comparable to our method, but a significantly lower \visual{} tidiness rate ($30.1\%$), thereby corroborating \hypo{2} about the necessity of an image-based tidiness-aware critic for grounding \semantically{} tidy actions.

\textbf{How do the \semantic{} and \visual{} tidiness rates vary with increasing complexity of the table layouts?} To investigate this, we subdivided the $70$ cases into $7$ subsets based on task complexity, and plotted the tidiness rates of each subset in Fig. \ref{fig:in_distribution_barplot} (b). Task complexity was controlled by varying i) the diversity of object types, ii) the quantity of objects, and iii) the requirement for 3D reasoning. Each marker in Fig. \ref{fig:in_distribution_barplot} (b) represents a distinct combination of these three factors. markers closer to the top-right corner of the plot correlate to higher average tidiness in the final layouts.

\textbf{Our approach} consistently achieves high rates for both \semantic{} and \visual{} tidiness, as demonstrated by the tight clustering of blue markers in the top right corner of Fig. \ref{fig:in_distribution_barplot}. Intuitively, increasing the table layout complexity presents challenges for both \semantic{} and \visual{} tidiness: reasoning over \emph{implicit} object features and functionalities becomes necessary for meaningful clustering, while capturing the nuances of \visual{} tidiness becomes more difficult due to the combinatorial variations in object arrangements. These results empirically validate the effectiveness of our approach. By decomposing the task into \semantic{} and \visual{} tidiness components and training the tidiness score using the 2-stage random walks, our approach effectively manages the task complexity and encodes sufficient discriminative power for \visual{} tidiness.

\textbf{The end-to-end LLM (baseline 1)} exhibits a decrease in both \semantic{} and \visual{} tidiness rates as task layouts become more complex. While baseline 1 shows some 2D spatial reasoning capability for simple table layouts (as shown in Fig. \ref{fig:tidy_instance} (a)), the rates decline as the number of object types increases, as shown by the darker markers closer to the left-bottom corner. This decline can be attributed to the simultaneous reasoning required for both \semantic{} and \visual{} tidiness, which proves challenging to scale for complex layouts. Moreover, the larger orange markers consistently fall below the smaller ones, indicating that an increase in object numbers has a negative impact on the \visual{} tidiness rate. This deterioration stems from the LLM generating physically infeasible actions, likely due to its limited capacity in spatial reasoning. This limited capacity is further evidenced by the low rates for layouts requiring 3D reasoning, depicted by the clustering of orange triangle marks in the far left, bottom corner.

\textbf{The LLM + collision checking (baseline 2)} achieves comparable \semantic{} tidiness rates to our approach, while its \visual{} tidiness rates remain unsatisfactory. This supports \hypo{2} that incorporating visual-spatial cues is essential for \visual{} tidiness. Despite the poor \visual{} tidiness rate, baseline 2 scores high on \semantic{} tidiness across all scenes. We attribute this to the task decomposition, where the LLM-based policy generator exclusively focuses on the semantic object grouping. Interestingly, although baseline 2 and our approach share the same object-centric policies, our approach outperforms baseline 2 in \semantic{} tidiness rates. This is because our evaluation is based on the inspection of the final layout image, and poor \visual{} tidiness can make object groupings appear unclear, even if they are reasonable. This highlights the importance of \visual{} tidiness beyond aesthetics; a \visually{} tidy layout enhances the clarity of object grouping, making it easier to find and retrieve objects.

\subsection{Generalization}

\begin{figure}[t]
    \centering
     \centering
     \includegraphics[width=1\linewidth]{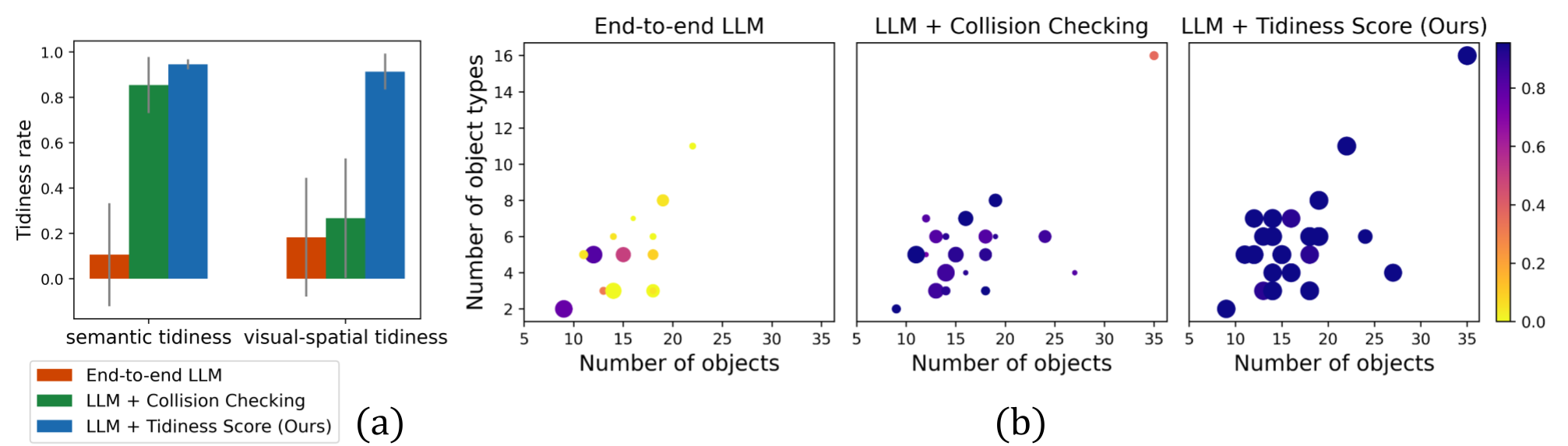}
    \caption{\small \textbf{Generalization Performance.} The left bar chart shows the mean \semantic{} and \visual{} tidiness scores for 30 test cases involving novel objects, with our method (blue bars) demonstrating high performance. The left scatter plot visualizes the correlation between the increase in object variety and quantity, and tidiness rates. Each marker represents a test case, where marker size and color represent \visual{} and \semantic{} tidiness scores, respectively. Darker and larger marker indicate higher \semantic{} and \visual{} tidiness rates, respectively. As the markers approach the top right corner, the complexity of the layouts increases.
    Our method's markers are notably darker and larger, indicating consistently superior \semantic{} and \visual{} tidiness performance.}
    \label{fig:out_dist_plot}
\vspace{-18pt}
\end{figure}

We evaluate the generalizability of our approach by tidying tables with unseen objects in 30 different table layouts, as illustrated in Fig. \ref{fig:objects} (\subfig b,\subfig c). These test cases include highly complex layouts with a large number of objects from a variety of types within the same scene. We present the tidiness rates for each method in Fig. \ref{fig:out_dist_plot} (a).

Our method achieves both \semantic{} and \visual{} tidiness for the majority of cases. This demonstrates the generalization capability of our approach, particularly the learned image-based tidiness score, which can handle unseen objects with varying shapes and sizes. Baseline 2 achieves high rates for \semantic{} tidiness but low scores for \visual{} tidiness, while baseline 1 scores poorly in both rates. Interestingly, baseline 1 achieves higher \visual{} tidiness rates than its \semantic{} tidiness: it occasionally arranges objects with similar sizes neatly in a line, disregarding their types and functionalities. This outcome can be attributed to the lack of an explicit structure in the end-to-end LLM approach for querying \semantic{} tidiness, which underscores the significance of task decomposition in our approach.

The scatter plot in Fig. \ref{fig:out_dist_plot} (b) shows how the \semantic{} and \visual{} tidiness rates change with increasing numbers and types of objects for each method. For baseline 1, both \semantic{} and \visual{} tidiness rates decline significantly as the layout complexity increases. For baseline 2, as the layouts become more complex, the markers slightly drop in size while reduce more visibly in the color intensity, indicating a consistently high \semantic{} tidiness rate but a decreasing \visual{} tidiness rate. Conversely, our proposed method consistently yield large and dark markers, suggesting that our table tidying capability remains robust even with novel objects and increasingly complex layouts.

We show case our method on a complex table setup in Fig. \ref{fig:tidy_instance} (h): the objects have various types but few objects per type. Both \semantic{} and \visual{} tidiness are challenging here. For \semantic{} tidiness, exact type-based grouping isn't ideal, as it results in scattered objects all over the table. Instead, we must reason about the implicit categories and similarities to meaningfully cluster the objects. \visual{} tidiness is also difficult due to the mix of types and shapes within clusters. We achieves both \semantic{} and \visual{} tidiness in this complex scene: we cluster objects as preserved food and snacks and adjust for varying shapes within each cluster. Taking the arrangement of the food cans for instance, tomato cans are neatly lined up (green box), tuna cans are stacked and aligned with tomato cans (yellow box), and the meat can fills remaining the empty corner (blue box), creating a compact and tidy block. This showcases our adaptability in fusing different notions of \visual{} tidiness based on the object shapes.

\section{Discussion} 
\label{sec:conclusion}

In this work, we ground the common-sense concept of \textit{tidiness} and propose an approach for tabletop object re-arrangement. Although LLMs capture \semantic{} tidiness,  visual nuances exceeds their capability. To address this, we learn a compact image-based critic using naturally synthesized data to complement the LLM's policy proposal. Our approach effectively demonstrates table tidying with complex and unseen object combinations.

One limitation of our current work is our 2D top-down scene representation, which restricts  complex 3D reasoning.
Currently, our 3D reasoning only involves the stacking of similarly-sized objects; such preference can be captured by our 2D model by favoring the layout with objects ``disappearing'' from the 2D image. Grounding the concept of \emph{tidiness} in 3D requires 3D object representation, scene conceptualization, and reasoning over other factors contributing to \textit{tidiness}, such as physical stability. Moreover, achieving a fine-grained 3D configuration between objects may require sophisticated motion planning. 

Another limitation is the assumed relatively uncluttered table settings. Substantially cluttered settings present significant challenges due to partial visibility and complex task planning. For instance, placing one object near another requires decluttering to create more space first.

These limitations suggest potential future research direction. Integrating 3D scene and object representation, belief tracking, and motion planning can provide an in-depth grounding of \emph{tidiness}, especially in complex scenarios involving 3D reasoning, occlusion, and cluttering.

\bibliographystyle{IEEEtran}
\bibliography{IEEEexample}

\end{document}